\documentclass[conference]{IEEEtran}
\IEEEoverridecommandlockouts
\usepackage{cite}
\usepackage{amsmath,amssymb,amsfonts}
\usepackage{algorithmic}
\usepackage{graphicx}
\usepackage{textcomp}
\usepackage{xcolor}
\usepackage{hyperref}
\usepackage{booktabs}
\usepackage{multirow}
\def\BibTeX{{\rm B\kern-.05em{\sc i\kern-.025em b}\kern-.08em
    T\kern-.1667em\lower.7ex\hbox{E}\kern-.125emX}}
\begin{document}

\title{Effectiveness of Counter-Speech against Abusive Content: A Multidimensional Annotation and Classification Study\\
}

\author{\IEEEauthorblockN{Greta Damo}
\IEEEauthorblockA{\textit{Université Côte d’Azur} \\
\textit{Inria, CNRS, I3S}\\
Sophia Antipolis, France \\
greta.damo@univ-cotedazur.fr}
\and
\IEEEauthorblockN{Elena Cabrio}
\IEEEauthorblockA{\textit{Université Côte d’Azur} \\
\textit{Inria, CNRS, I3S}\\
Sophia Antipolis, France \\
elena.cabrio@univ-cotedazur.fr}
\and
\IEEEauthorblockN{Serena Villata}
\IEEEauthorblockA{\textit{Université Côte d’Azur} \\
\textit{Inria, CNRS, I3S}\\
Sophia Antipolis, France \\
serena.villata@univ-cotedazur.fr}
}

\maketitle

\begin{abstract}
Counter-speech (CS) is a key strategy for mitigating online Hate Speech (HS), yet defining the criteria to assess its effectiveness remains an open challenge. We propose a novel computational framework for CS effectiveness classification, grounded in linguistics, communication and argumentation concepts. Our framework defines six core dimensions -- Clarity, Evidence, Emotional Appeal, Rebuttal, Audience Adaptation, and Fairness -- which we use to annotate 4,214 CS instances from two benchmark datasets, resulting in a novel linguistic resource released to the community. In addition, we propose two classification strategies, multi-task and dependency-based, achieving strong results (0.94 and 0.96 average F1 respectively on both expert- and user-written CS), outperforming standard baselines, and revealing strong interdependence among dimensions.
\end{abstract}

\noindent{\color{red}\textbf{Content warning}: this paper contains unobfuscated examples some readers may find offensive.}


\section{Introduction}
The rise of social media has intensified concerns over online Hate Speech (HS). 
Counter-speech (CS) has emerged as an effective strategy for mitigating harmful online discourse. It involves non-aggressive responses, that use credible evidence, factual arguments, and alternative perspectives to challenge hate speech \cite{schieb2016governing, benesch2014countering}. Counter-speech has been studied and approached through various lenses: some see it as direct responses to hate speech \cite{mathew2019thou, ashida-komachi-2022-towards}, while others argue it should condemn hate, challenge and offer alternative viewpoints, and offer support to the victim \cite{vidgen2021introducing, hangartner2021empathy, vidgen2020detecting, He2021}.
CS is also valued for its potential to shift beliefs and reduce future hate~\cite{qian-etal-2019-benchmark, Rieger2018}. CS can be written by experts, online users, or chatbots (based on Large Language Models -- LLMs). Expert-written CS typically presents structured reasoning and appeals to shared human values,
whereas user- and LLM-generated responses, in particular, tend to be more generic and emotional, and often lacks depth and relies on surface-level rebuttals \cite{mun-etal-2023-beyond, tekiroglu-etal-2022-using, tekiroglu-etal-2020-generating}.

\begin{figure}[t] \small \centering \includegraphics[width=0.48\textwidth]{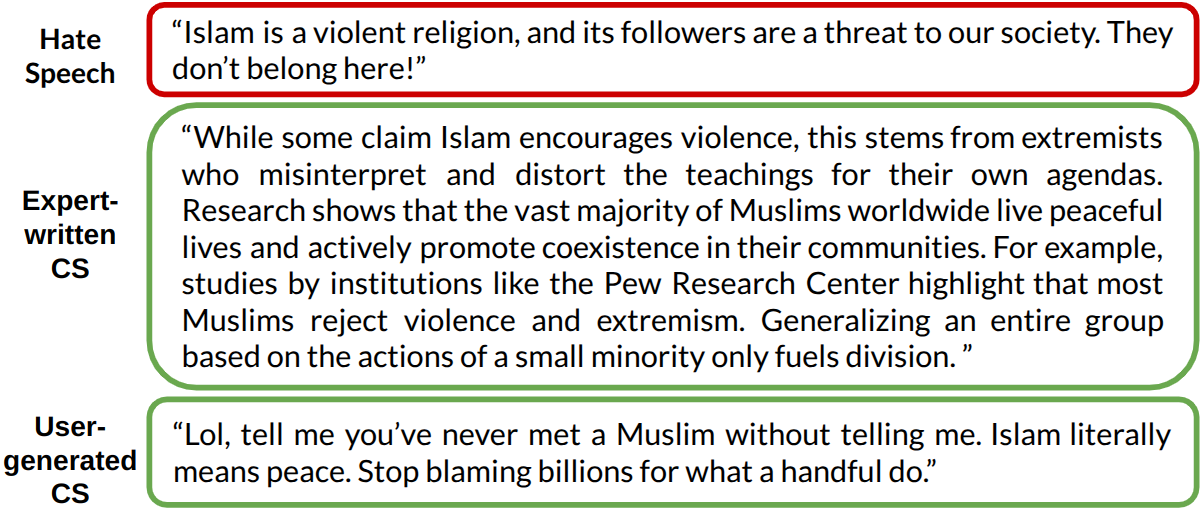} \caption{Examples of expert-written and user-written counter-speech.} \label{fig:intro} \end{figure}

This study evaluates expert- and user-written CS effectiveness using two benchmark datasets \cite{albanyan2022pinpointing, chung-etal-2019-conan}.This allows us to compare how the origin of a counter-speech message influences its effectiveness. 
Figure~\ref{fig:intro} shows an example of a hateful message with user- and expert-written CS responses.
Current evaluations of counter-speech effectiveness rely on surface-level metrics -- coherence, relevance, or non-offensiveness -- assuming well-formed CS is 
inherently effective \cite{chung-etal-2021-towards, zhu-bhat-2021-generate, baheti-etal-2021-just, ashida-komachi-2022-towards, mun-etal-2023-beyond}. But concentrating on CS surface form only is not enough to achieve shifting attitudes or reducing harm. 
Assessing CS effectiveness is crucial, as it has the potential for significant societal impact, by reducing online hostility and potentially curbing offline violence and discrimination. 

To address this gap, this study introduces a novel structured framework grounded in linguistics, communication, rhetoric, and argumentation theories, helping to evaluate counter-speech effectiveness. 
Our contribution is twofold: \textbf{(1) a new multi-dimensional framework for automatic counter-speech effectiveness classification} across six key dimensions -- Clarity, Evidence, Emotional Appeal, Rebuttal, Audience Adaptation, and Fairness -- from linguistics, communication, rhetoric and argumentation theories, and two classification strategies, \textbf{multi-task} and \textbf{dependency-based} modeling, that outperform standard baselines, achieving average F1 scores of 0.94 and 0.96 respectively, when combining user-written and expert-written CS data.
\textbf{(2) a novel human annotated resource} comprising two expert- \cite{chung-etal-2019-conan} and user-written \cite{albanyan2022pinpointing} counter-speech datasets, enriched with the annotation layer with the six dimensions of our effectiveness framework\footnote{Datasets, guidelines, and code are available at this \href{https://github.com/grexit-d/counter_speech_effectiveness}{link}.}. 

The remainder of the paper is structured as follows. Section \ref{sec:related} reviews prior studies on counter-speech and its evaluation. Section \ref{sec:background} defines the six dimensions used to assess CS effectiveness, tailoring them to existing linguistics and argumentation theories. Section \ref{sec:datasets} details the datasets used for the annotation and the classification task, and the annotation process together with the Inter-Annotator-Agreement results. Section \ref{sec:experiments} outlines the models used, and the experimental setup. Section \ref{sec:results} presents the results of our classification task, showing comparisons between the baselines and our proposed models. Section \ref{sec:error_discussion} identifies and discusses challenges in specific dimensions and limitations. Finally, Section \ref{sec:conclusion} summarizes key findings and future directions.

\section{Related Work}
\label{sec:related}

Research on counter-speech has gained increasing attention as a strategy for combating online hate speech. However, most studies have focused on hate speech detection and counter-speech generation, while the systematic evaluation of CS effectiveness remains under-explored. In this section, we review key contributions on counter-speech analysis, emphasizing the need for a comprehensive CS evaluation framework.

\subsection{Hate Speech \& Counter Speech Detection}

HS detection is a well-studied NLP task, initially tackled as binary classification on English datasets \cite{founta2018large, davidson2017automated}, later extended to fine-grained labels for discrimination and microaggressions \cite{guest2021expert, grimminger2021hate, ross2017measuring, sap2020social, hede2021toxicity, wiegand2021implicitly}. Early approaches relied on traditional Machine Learning \cite{waseem2016racist, aziz2021hate}, while recent methods leverage transformers like BERT for better contextual modeling \cite{vashistha2021online, khan2023transformer, wadud2023deepbert, pavlopoulos2017deeper}.

\noindent On the other hand, 
counter-speech, introduced as a mitigation strategy by \cite{benesch2016counterspeech}, has not yet been extensively explored from a detection standpoint. Among the studies, 
\cite{garland-etal-2020-countering} propose an early HS/CS detection pipeline, by applying an ensemble learning algorithm to classify HS and CS tweets, achieving F1 scores between 0.76 and 0.97, emphasizing counter-speech complexity due to subjectivity and definitional ambiguity. 

\subsection{Counter Speech Classification \& Evaluation} 

Concerning counter-speech classification, 
\cite{poudhar-etal-2024-strategy} find that a one-shot prompted LLM achieves promising accuracy in classifying manually labeled CS strategies.
\noindent \cite{Mathew2018} analyze HS/CS tweet pairs, using a boosting algorithm with TF-IDF and lexical features, which achieves 0.77 F1. \cite{mathew2019thou} also analyze CS and neutral comments on YouTube, finding that TF-IDF vectors combined with logistic regression achieve 0.73 F1. 
\cite{Wright2017} conducts a qualitative analysis of counter-speech examples, while
\cite{Ziems2020} analyze tweets during the COVID-19 crisis, achieving an F1 score of 0.49 for CS. 

A key gap in the literature is the lack of standardized evaluation methods for assessing
human- or machine-generated counter-speech. Several studies propose evaluation criteria, but no unified framework emerged. \cite{zheng-etal-2023-makes} assess CS effectiveness through human judgments, focusing on comfort for targets and bystander empathy. \cite{chung-etal-2021-towards} and \cite{zhu-bhat-2021-generate} similarly rely on human ratings based on qualities such as informativeness, coherence, and stance. \cite{ashida-komachi-2022-towards} evaluate CS suitability using offensiveness and stance, while \cite{jones-etal-2024-multi} propose an LLM-based evaluation framework, grounded in NGO guidelines, which shows strong alignment with human judgments. \cite{yu2024hate} introduce a conversation-level incivility metric, arguing that CS should be judged by its downstream impact on discourse. Finally, \cite{song-etal-2025-assessing} focus on evaluating the human-likeness of AI-generated counter-speech.

While prior work shows growing interest in CS evaluation, 
we advance beyond previous studies by introducing a human-annotated resource based on our framework that captures fine-grained dimensions of 
structural coherence, content strength, rhetorical impact, and linguistic strategies, to assess its effectiveness. We also develop classification methods that leverage this framework.

\section{Effectiveness Metrics}
\label{sec:background}


We propose a framework to evaluate counter-speech effectiveness across six dimensions: Clarity, Evidence, Emotional Appeal, Rebuttal, Audience Adaptation, and Fairness. 
These are grounded in 
communication theory.

\begin{itemize}

\item \textbf{Clarity} ensures structural coherence and logical flow. Effective counter-speech should be clear and logically structured, so that it is easier for the audience to follow the reasoning and understand the main points 
\cite{persing-ng-2013-modeling, wachsmuth-etal-2016-using, wachsmuth-etal-2017-computational, stapleton2015assessing}.
\item \textbf{Evidence} ensures that effective counter-speech is supported by relevant evidence and examples that support the claims and make it more compelling to the audience. They should be specific to the topics addressed in the hateful message. Studies indicate that presenting multiple pieces of evidence in an argument, such as statistics or expert testimony, enhances persuasiveness and effectiveness \cite{wachsmuth-etal-2017-computational, rinott-etal-2015-show, Rahimi2014, reinard1988empirical}.
\item \textbf{Emotional Appeal} is a rhetorical strategy, used to evoke empathy, or other emotions in the audience, which can help strengthen the counter-speech impact. The importance of emotions and empathy in designing effective counter-speech has been investigated by numerous studies \cite{wachsmuth-etal-2017-computational, Aristotle2007, Govier2010, hangartner2021empathy, EmotionMessages2022, Marone2015}.
\item \textbf{Rebuttal} shows how using arguments to anticipate and address potential counterarguments, objections, and opposing views improves credibility and strengthens persuasion \cite{wachsmuth-etal-2017-computational, Aristotle2007, Toulmin1958, Damer2009, Granger2009, okeefe1999handling, onoda2015highlighting}.
\item \textbf{Audience Adaptation} involves tailoring the language of the counter-speech to a specific audience, taking into account their level of linguistic ability and knowledge, thus improving understanding and relatability.
From a linguistic point of view, a counter-speech is more effective and understandable if the language employed is similar to the one spoken by the audience \cite{pickering2004toward, danescu2011mark, wang-etal-2014-model}.
\item \textbf{Fairness} promotes the use of appropriate and respectful language, respecting freedom of expression, without censoring or dehumanizing the opposing viewpoint.
Fair language supports ethical discourse and improves effectiveness \cite{cialdini1993influence, schreier1995thats}.
\end{itemize}

Unlike prior frameworks, focused narrowly on toxicity or engagement, our approach enables fine-grained analysis informed by the theoretical perspectives described in argumentation and communication studies. They are divided into binary and categorical metrics.
\begin{table}[t]
\caption{Annotation guidelines for CS effectiveness.}
\centering
\scriptsize
\renewcommand{\arraystretch}{1.05}
\setlength{\tabcolsep}{2pt}
\begin{tabular}{p{1.2cm} | p{6.2cm}}
\toprule
\multicolumn{2}{l}{\textbf{Categorical Dimensions (1–3 Likert Scale)}} \\
\midrule
\textbf{Clarity} & 
3: Clear, logically structured, directly addresses the HS topic. \newline
2: Mostly clear or slightly generic. \newline
1: Vague or incoherent. \\
\midrule
\textbf{Evidence} & 
3: Multiple pieces of information as supporting evidence. \newline
2: One supporting information. \newline
1: No information as evidence. \\
\midrule
\textbf{Rebuttal} & 
3: Multiple rebuttals targeting specific parts of the HS. \newline
2: One rebuttal. \newline
1: No rebuttal. \\
\midrule
\textbf{Fairness} & 
3: Respectful, no swearing, no attacks against the hater. \newline
2: Mostly polite, no attacks, mild offensive. \newline
1: Aggressive, including swearing and personal attacks. \\
\midrule
\multicolumn{2}{l}{\textbf{Binary Dimensions (0/1)}} \\
\midrule
\textbf{Emotional} \textbf{Appeal} & 
1: Language that evokes emotions (either positive or negative). \newline
0: Neutral tone and language. \\
\midrule
\textbf{Audience} \textbf{Adaptation} & 
1: Matches HS tone/complexity. \newline
0: Mismatched HS tone or complexity. \\
\bottomrule
\end{tabular}
\label{tab:cs_dimensions}
\end{table}

\noindent \textbf{Binary metrics.} Emotional Appeal and Audience Adaptation are binary (0-1), where 1 indicates presence and 0 absence of the dimension. Due to their inherent subjectiveness and context-dependency, a binary classification (presence vs. absence) seems reasonable. 
Similar to prior work simplifying subjective dimensions into binary formats \cite{habernal2016which, wachsmuth-etal-2017-computational, oraby2017thats, koszowy2024pathos}, we adopt this approach to reduce ambiguity and subjectivity.

\noindent \textbf{Categorical metrics.} Clarity, Evidence, Rebuttal, and Fairness are categorical dimensions, using a 1–3 Likert scale, with 3 as the best score. Compared to the binary variables, these dimensions rely on more objective indicators (for example, the number of pieces of evidence), therefore, a 1-3 scale allows annotators to capture gradual differences, distinguishing between weak, moderate, and strong instances. 
Table \ref{tab:cs_dimensions} reports a summary of the definitions for the annotation of the effectiveness dimensions.

\section{Datasets Description}
\label{sec:datasets}

We extend the annotation on two existing datasets: \textbf{CONAN} \cite{chung-etal-2019-conan} and the \textbf{Twitter Dataset} \cite{albanyan2022pinpointing}. We select CONAN as it is the first expert-curated dataset of HS/CS pairs, widely used as a benchmark. It is a multilingual (English, French, Italian) dataset centered on Islamophobia, containing 4,078 expert-annotated HS/CS pairs. Through translation and paraphrasing, it is expanded to 14,988 pairs. For our experiments, we retain only the English, non-augmented instances, resulting in 3,847 pairs. The Twitter Dataset\footnote{Only tweet IDs are publicly available due to privacy policies. We obtained the full data upon request from the authors.} 
is a real-world dataset, containing 5,652 hateful tweets and replies obtained from social media (Twitter/X), capturing the brevity and style typical of online discourse. 
We focus on the subset labeled as counter-speech, where a clear target is identifiable, yielding 367 HS/CS pairs.
Together, these datasets form a combined benchmark of 4,214 HS/CS pairs, used in our experiments.





\noindent \textbf{Annotation \& IAA.} We define annotation guidelines to label CS effectiveness dimensions. 
To consolidate these guidelines\footnote{Complete guidelines are available at this \hyperref[https://anonymous.4open.science/r/counter-speech_effectiveness-7DFC/README.md]{link}}, a pilot study was conducted on 50 HS/CS pairs from each dataset, involving three annotators \footnote{Two female, one other; age group: 21–30; education level: PhD students.} with background in computer science and linguistics.
Through different reconciliation phases, the guidelines were improved iteratively.
After an initial round, annotators reviewed disagreements to resolve confusion or misinterpretation. Based on their feedback, the guidelines were updated, and a second and final annotation round was conducted.
To measure inter-annotator agreement (IAA), Fleiss’s $\kappa$ was used for binary dimensions, Krippendorff's $\alpha$ for categorical ones, and Percent Agreement for Audience Adaptation due to near-perfect consensus (as other metrics underestimate high non-random agreement). Table~\ref{tab:iaa-eff_scores_avg} shows IAA results among the three annotators. The IAA of the second round of annotations improved significantly, compared to the first one, as the annotators had the possibility to discuss and refine their understanding of the guidelines. 
Overall, the final IAA scores
indicate strong agreement across all dimensions, so the remaining part of the dataset was labeled by one of the annotators.
Table~\ref{tab:iaa-eff_scores_avg} also shows average effectiveness scores across both datasets. Expert-written counter-speech obtains an average effectiveness score of 1.62, demonstrating its higher effectiveness compared to user-written counter-speech with an average score of 1.34. CONAN scores higher in all dimensions except emotional appeal, suggesting that online users rely more on emotions than evidential reasoning or facts.
%
\begin{table}[h]
\caption{\footnotesize IAA from two annotation rounds (Fleiss’ $\kappa$ for binary variables, and Krippendorff’s $\alpha$ for categorical ones) and average effectiveness scores.}
\centering
\scriptsize
\renewcommand{\arraystretch}{1.05}
\resizebox{0.48\textwidth}{!}{
\begin{tabular}{lcccccc}
\toprule
\multirow{2}{*}{\textbf{Dimension}} 
& \multicolumn{3}{c}{\textbf{CONAN}} 
& \multicolumn{3}{c}{\textbf{Twitter}} \\
& 1st & 2nd & avg. 
& 1st & 2nd & avg. \\
\midrule
\textit{Emotional}        & 0.30 & 0.65 & 0.29  & 0.40 & 0.62 & 0.45 \\
\textit{Audience}         & 1.00 & 1.00 & 0.99  & 0.99 & 0.99 & 1.00 \\
\textit{Clarity}          & 0.29 & 0.75 & 2.72  & 0.41 & 0.82 & 1.96 \\
\textit{Evidence}         & 0.27 & 0.73 & 1.47  & 0.34 & 0.75 & 1.07 \\
\textit{Rebuttal}         & 0.21 & 0.78 & 1.31  & 0.42 & 0.62 & 1.18 \\
\textit{Fairness}         & 0.63 & 0.79   & 2.95  & 0.53 & 0.94 & 2.35 \\
\midrule
\textbf{Total AVG}        & --   & --   & \textbf{1.62} & -- & -- & \textbf{1.34} \\
\bottomrule
\end{tabular}
}
\label{tab:iaa-eff_scores_avg}
\end{table}

\section{Experimental Setting}
\label{sec:experiments}

In this section, we describe the model configurations used in our experiments. The goal is to predict CS effectiveness scores across six dimensions. We use BERT classifiers based on \texttt{bert-base-uncased} \cite{bert} to predict the effectiveness scores for a given counter-speech response. 
In all configurations, the \texttt{max\_length} is set to 128. The \texttt{batch\_size} is 16, and the \texttt{learning\_rate} is 2e-5. We run the experiments using five different seeds (0, 1, 2, 3, 42), and we average the results. 
To perform the experiments, we concatenate both CONAN and Twitter datasets, and then we randomly split it into train, validation, and test subsets, with a percentage of 70-10-20. We also perform cross-validation, by using a single dataset for training and validation, and the other for testing. The experiments were run on one A100 GPU.

We use BERT as the sole model to focus on improving it through our framework. By not comparing across different architectures, we isolate the effect of our proposed models, ensuring that observed gains result from our contributions, rather than differences between model architectures.

We assess the following configurations:

\begin{itemize}

\item \textbf{Bert\_CS}: counter-speech text embeddings are used as input. Binary dimensions are trained using the Binary Cross-Entropy (BCE) loss function, and categorical ones with Cross-Entropy Loss (CE).
\item \textbf{Bert\_CS\_HS}: identical to Bert\_CS, but the input is the concatenation of hate speech and counter-speech text embeddings, to add more context.
\item \textbf{Multi-task\_divided.} Binary dimensions are trained jointly using BCE (for \textit{emotional\_appeal}) and Focal Loss (to mitigate class imbalance for \textit{audience\_adaptation}), while categorical dimensions are trained separately using CE. Final predictions are obtained by combining all loss components. This setting assumes partial correlation among grouped dimensions.
\item \textbf{Multi-task\_united.} All dimensions (binary and categorical) are trained together using their respective loss functions in a single multi-task model. The total loss is computed as the sum of individual losses. This configuration assumes stronger interdependence among all dimensions. 
\item \textbf{Dependency\_matrix\_3e.} All the dimensions are considered together, with their respective loss function (BCE for \textit{emotional\_appeal}, Focal Loss for \textit{audience\_adaptation}, and Cross Entropy Loss for all the multi-label dimensions), which are summed up at the end. Additionally, there is a new parameter, a \texttt{learnable dependency matrix}, to capture pairwise relationships between dimensions. Its rows and columns contain values between 0 and 1, that are weights accounting for the dependency between pairs of dimensions. Its weights are randomly initialized, and their best value is learned during the training phase for 3 epochs.
\item \textbf{Dependency\_matrix\_6e} has the same configuration of Dependency\_matrix\_3e, but it is trained for 6 epochs to assess the effect of extended training.
\end{itemize}

\noindent We experiment with two types of input embeddings: (i) concatenated Hate Speech (HS) and Counter Speech (CS) embeddings to provide broader context and (ii) CS embeddings alone. The latter consistently yields slightly better performance. Consequently, we adopt CS-only embeddings for all our proposed configurations (\textit{Multi-task\_divided}, \textit{Multi-task\_united}, \textit{Dependency\_matrix\_3e}, and \textit{Dependency\_matrix\_6e}). 

\section{Results}
\label{sec:results}

\begin{table*}[t]
\caption{\footnotesize F1 scores (± std) for all models. Statistically significant results are marked with *. AVG: macro-average. Emo.: Emotional Appeal, Aud.: Audience Adaptation, Clarity: Clarity, Evid.: Evidence, Rebut.: Rebuttal, Fair.: Fairness.}
\centering
\large
\renewcommand{\arraystretch}{1.8}
\setlength{\tabcolsep}{2.8pt}
\resizebox{\textwidth}{!}{%
\begin{tabular}{@{}lccccccc|ccccccc|ccccccc@{}}
\toprule
\textbf{Model} 
& \multicolumn{7}{c|}{\textbf{Train: Combined Dataset}} 
& \multicolumn{7}{c|}{\textbf{Train: Twitter $\rightarrow$ CONAN}} 
& \multicolumn{7}{c}{\textbf{Train: CONAN $\rightarrow$ Twitter}} \\
\cmidrule(r){2-8} \cmidrule(r){9-15} \cmidrule(r){16-22}
& \textbf{Emo.} & \textbf{Aud.} & \textbf{Clarity} & \textbf{Evid.} & \textbf{Rebut.} & \textbf{Fair.} & \textbf{AVG} 
& \textbf{Emo.} & \textbf{Aud.} & \textbf{Clarity} & \textbf{Evid.} & \textbf{Rebut.} & \textbf{Fair.} & \textbf{AVG} 
& \textbf{Emo.} & \textbf{Aud.} & \textbf{Clarity} & \textbf{Evid.} & \textbf{Rebut.} & \textbf{Fair.} & \textbf{AVG} \\
\midrule
\textit{bert\_cs}       
& 0.67\textsuperscript{$\pm$ 0.09} & 0.99\textsuperscript{$\pm$ 0.00} & 0.94\textsuperscript{$\pm$ 0.01} & 0.96\textsuperscript{$\pm$ 0.01} & 0.96\textsuperscript{$\pm$ 0.01} & 0.95\textsuperscript{$\pm$ 0.01} & 0.91\textsuperscript{$\pm$ 0.02} 
& 0.14\textsuperscript{$\pm$ 0.00} & 0.99\textsuperscript{$\pm$ 0.00} & 0.74\textsuperscript{$\pm$ 0.05} & 0.45\textsuperscript{$\pm$ 0.06} & 0.55\textsuperscript{$\pm$ 0.11} & 0.92\textsuperscript{$\pm$ 0.01} & 0.63\textsuperscript{$\pm$ 0.03} 
& 0.30\textsuperscript{$\pm$ 0.01} & 1.00\textsuperscript{$\pm$ 0.00} & 0.41\textsuperscript{$\pm$ 0.04} & 0.86\textsuperscript{$\pm$ 0.09} & 0.77\textsuperscript{$\pm$ 0.02} & 0.54\textsuperscript{$\pm$ 0.21} & 0.65\textsuperscript{$\pm$ 0.05} \\
\textit{bert\_cs\_hs}   
& 0.62\textsuperscript{$\pm$ 0.12} & 0.99\textsuperscript{$\pm$ 0.00} & 0.94\textsuperscript{$\pm$ 0.02} & 0.96\textsuperscript{$\pm$ 0.01} & 0.95\textsuperscript{$\pm$ 0.01} & 0.94\textsuperscript{$\pm$ 0.01} & 0.90\textsuperscript{$\pm$ 0.02} 
& 0.14\textsuperscript{$\pm$ 0.01} & 0.99\textsuperscript{$\pm$ 0.00} & 0.76\textsuperscript{$\pm$ 0.06} & 0.43\textsuperscript{$\pm$ 0.02} & 0.55\textsuperscript{$\pm$ 0.14} & 0.92\textsuperscript{$\pm$ 0.02} & 0.63\textsuperscript{$\pm$ 0.03} 
& 0.30\textsuperscript{$\pm$ 0.04} & 1.00\textsuperscript{$\pm$ 0.00} & 0.33\textsuperscript{$\pm$ 0.03} & 0.84\textsuperscript{$\pm$ 0.07} & 0.73\textsuperscript{$\pm$ 0.05} & 0.41\textsuperscript{$\pm$ 0.07} & 0.60\textsuperscript{$\pm$ 0.02} \\
\textit{multi-task\_d}  
& 0.92\textsuperscript{$\pm$ 0.02} & 0.99\textsuperscript{$\pm$ 0.00} & 0.92\textsuperscript{$\pm$ 0.01} & 0.93\textsuperscript{$\pm$ 0.01} & 0.94\textsuperscript{$\pm$ 0.03} & 0.92\textsuperscript{$\pm$ 0.03} & 0.94\textsuperscript{$\pm$ 0.01} 
& 0.55\textsuperscript{$\pm$ 0.17} & 0.99\textsuperscript{$\pm$ 0.00} & 0.51\textsuperscript{$\pm$ 0.24} & 0.44\textsuperscript{$\pm$ 0.01} & 0.60\textsuperscript{$\pm$ 0.05} & 0.89\textsuperscript{$\pm$ 0.13} & \textbf{0.66*}\textsuperscript{$\pm$ 0.04} 
& 0.49\textsuperscript{$\pm$ 0.07} & 1.00\textsuperscript{$\pm$ 0.00} & 0.42\textsuperscript{$\pm$ 0.02} & 0.91\textsuperscript{$\pm$ 0.04} & 0.80\textsuperscript{$\pm$ 0.02} & 0.32\textsuperscript{$\pm$ 0.03} & 0.65\textsuperscript{$\pm$ 0.02} \\
\textit{multi-task\_u}  
& 0.84\textsuperscript{$\pm$ 0.04} & 0.99\textsuperscript{$\pm$ 0.00} & 0.93\textsuperscript{$\pm$ 0.02} & 0.92\textsuperscript{$\pm$ 0.02} & 0.93\textsuperscript{$\pm$ 0.03} & 0.93\textsuperscript{$\pm$ 0.02} & 0.92\textsuperscript{$\pm$ 0.01} 
& 0.47\textsuperscript{$\pm$ 0.23} & 0.99\textsuperscript{$\pm$ 0.00} & 0.38\textsuperscript{$\pm$ 0.27} & 0.43\textsuperscript{$\pm$ 0.02} & 0.56\textsuperscript{$\pm$ 0.07} & 0.72\textsuperscript{$\pm$ 0.39} & 0.59\textsuperscript{$\pm$ 0.09} 
& 0.52\textsuperscript{$\pm$ 0.06} & 1.00\textsuperscript{$\pm$ 0.00} & 0.42\textsuperscript{$\pm$ 0.03} & 0.93\textsuperscript{$\pm$ 0.04} & 0.77\textsuperscript{$\pm$ 0.02} & 0.35\textsuperscript{$\pm$ 0.05} & 0.66\textsuperscript{$\pm$ 0.01} \\
\textit{dep.\_m.\_3e}   
& 0.80\textsuperscript{$\pm$ 0.10} & 0.99\textsuperscript{$\pm$ 0.00} & 0.92\textsuperscript{$\pm$ 0.01} & 0.90\textsuperscript{$\pm$ 0.05} & 0.91\textsuperscript{$\pm$ 0.03} & 0.93\textsuperscript{$\pm$ 0.03} & 0.91\textsuperscript{$\pm$ 0.02} 
& 0.40\textsuperscript{$\pm$ 0.24} & 0.99\textsuperscript{$\pm$ 0.00} & 0.48\textsuperscript{$\pm$ 0.22} & 0.43\textsuperscript{$\pm$ 0.03} & 0.53\textsuperscript{$\pm$ 0.09} & 0.70\textsuperscript{$\pm$ 0.13} & 0.59\textsuperscript{$\pm$ 0.08} 
& 0.54\textsuperscript{$\pm$ 0.02} & 1.00\textsuperscript{$\pm$ 0.00} & 0.41\textsuperscript{$\pm$ 0.02} & 0.92\textsuperscript{$\pm$ 0.02} & 0.77\textsuperscript{$\pm$ 0.02} & 0.34\textsuperscript{$\pm$ 0.05} & 0.66\textsuperscript{$\pm$ 0.01} \\
\textit{dep.\_m.\_6e}   
& 0.92\textsuperscript{$\pm$ 0.04} & 0.99\textsuperscript{$\pm$ 0.00} & 0.95\textsuperscript{$\pm$ 0.01} & 0.97\textsuperscript{$\pm$ 0.02} & 0.96\textsuperscript{$\pm$ 0.01} & 0.95\textsuperscript{$\pm$ 0.02} & \textbf{0.96*}\textsuperscript{$\pm$ 0.01} 
& 0.40\textsuperscript{$\pm$ 0.21} & 0.99\textsuperscript{$\pm$ 0.00} & 0.75\textsuperscript{$\pm$ 0.04} & 0.53\textsuperscript{$\pm$ 0.07} & 0.47\textsuperscript{$\pm$ 0.21} & 0.78\textsuperscript{$\pm$ 0.14} & 0.65\textsuperscript{$\pm$ 0.07} 
& 0.50\textsuperscript{$\pm$ 0.05} & 1.00\textsuperscript{$\pm$ 0.00} & 0.42\textsuperscript{$\pm$ 0.01} & 0.92\textsuperscript{$\pm$ 0.03} & 0.77\textsuperscript{$\pm$ 0.02} & 0.38\textsuperscript{$\pm$ 0.01} & \textbf{0.66*}\textsuperscript{$\pm$ 0.01} \\
\bottomrule
\end{tabular}%
}
\label{tab:combined_results}
\end{table*}

Table~\ref{tab:combined_results} shows average F1 scores across five runs for all model configurations across our counter-speech effectiveness dimensions. We compare BERT baselines, multi-task learning models, and dependency-based architectures (with CS embeddings). 
The dependency-based model \textit{Dependency\_matrix\_6e} consistently achieves the highest F1 score (0.96), outperforming all the others. This performance gain is likely due to the use of a learnable dependency matrix that captures the interrelations between dimensions, allowing the model to exploit cross-dimensional dependencies, unlike models that treat the six dimensions independently. 
Statistical tests confirm that the results are significantly different. A one-way ANOVA test, conducted on the average F1 scores across models, confirms a statistically significant difference in performance (\textit{F} = 12.92, \textit{p} $<$ 0.001), especially between the best model, \textit{Dependency\_matrix\_6e}, and all others (pairwise t-tests between it and each of the other models show statistically significant differences at the 1\% significance level).

\begin{figure}[t]
    \includegraphics[width=0.45\textwidth]{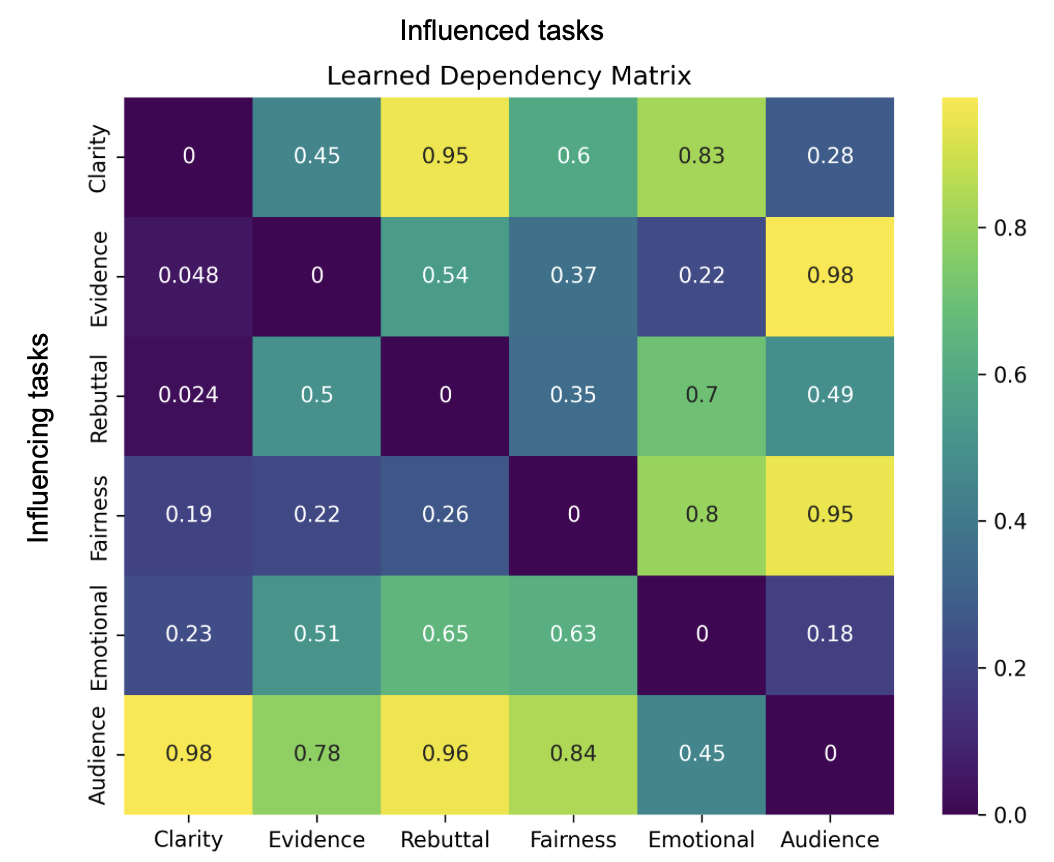}
    \caption{\footnotesize Learned Dependency Matrix obtained from \textit{Dependency\_matrix\_6e} on the combined dataset with seed 42.}
   \label{fig:dependency_matrix}
\end{figure}

Figure~\ref{fig:dependency_matrix} shows the dependency matrix learned by \textit{Dependency\_matrix\_6e}. Rows represent influenced tasks, columns represent influencing tasks, with higher values indicating stronger influence (diagonal values are zero as dimensions do not influence themselves).
We can see that Emotional Appeal is influenced by Fairness (0.63) and Rebuttal (0.65), while Audience Adaptation is shaped by Clarity (0.98), Rebuttal (0.96), and Fairness (0.84). Clarity is strongly influenced by Rebuttal (0.95) and Emotional Appeal (0.83); Evidence is influenced by Audience Adaptation (0.98); Fairness is influenced by Emotional Appeal (0.80) and Audience Adaptation (0.95), and
Rebuttal by Emotional Appeal (0.7).
These inter-dependencies help the model to improve its classification performance across dimensions.
Both multi-task models, \textit{Multi-task\_divided} (0.94) and \textit{Multi-task\_united} (0.92), also outperform the BERT baselines, suggesting that the multi-task framework is more effective, 
with only a slight performance difference between united and divided. 
Fine-tuned BERT models remain competitive: \textit{Bert\_CS} (0.91) slightly outperforms \textit{Bert\_CS\_HS} (0.90). This suggests that incorporating HS data does not significantly improve classification performance.

\noindent Emotional Appeal remains the hardest dimension to classify, showing the most variation across models, ranging from 0.67 to 0.92 F1 for the combined dataset. In this dimension, dependency-based and multi-task models perform best (up to 0.92 for \textit{Dependency\_matrix\_6e} and \textit{Multi-task\_divided}), while BERT baselines lag behind (0.67 and 0.62),
suggesting that richer architectures better capture emotional cues. Audience Adaptation scores above 0.99 in all configurations. Although this can be due to class imbalance, we address this concern using focal loss during training, which focuses the training on hard, misclassified examples and reduces the impact of easy, well-classified examples.
By contrast, Clarity and Evidence show greater variation, pointing to their higher classification difficulty.

\noindent \textbf{Cross-Domain Generalization.} 
To assess model generalization, we perform cross-domain evaluation. 
In the first setting, models are trained on CONAN (expert-written CS) and tested on Twitter Dataset (user-generated CS).
All models show a marked performance drop, with \textit{Dependency\_matrix\_6e} performing best (0.66) alongside \textit{Dependency\_matrix\_3e} and \textit{Multi-task\_united}.
Emotional Appeal declines the most, 
suggesting a domain shift in how emotional content is conveyed. Clarity, Fairness, and Rebuttal scores are also low, reflecting the less structured nature of user-written CS. 
In the reverse setting, where models are trained on Twitter Dataset (user-written CS) and tested on CONAN (expert-generated CS),
results are similar. \textit{Multi-task\_divided} achieves the highest F1 score (0.66), followed closely by \textit{Dependency\_matrix\_6e} (0.65), which remains highly competitive.
The remaining models obtain lower average F1 scores, indicating less robust generalization in this transfer setup. Models trained on expert-written CS (Twitter $\rightarrow$ CONAN) 
seem more adaptable, but Emotional Appeal remains challenging across domains, ranging from 0.14 to 0.55. In contrast, Audience Adaptation remains consistently high, showing robustness across domains, while Fairness remains strong only on expert-written data.
Overall, our models generalize reasonably well, achieving a good F1 score in these cross-dataset tests. This suggests that models learn meaningful, transferable patterns beyond dataset-specific characteristics.
Given the variability in emotional appeal, domain adaptation strategies or dataset augmentation could help models better capture diverse emotional expressions across different counter-speech sources.

\begin{table*}[t]
\caption{F1 scores per effectiveness dimension on Twitter and CONAN datasets. AVG: macro-average. Emo.: Emotional Appeal, Aud.: Audience Adaptation, Clarity: Clarity, Evid.: Evidence, Rebut.: Rebuttal, Fair.: Fairness.}
\centering
\setlength{\tabcolsep}{2.5pt}
\large
\renewcommand{\arraystretch}{1.6}
\resizebox{\textwidth}{!}{
\begin{tabular}{lccccccc|ccccccc}
\toprule
\multirow{2}{*}{\textbf{Model}} & \multicolumn{7}{c|}{\textbf{Twitter Dataset}} & \multicolumn{7}{c}{\textbf{CONAN Dataset}} \\
 & \textbf{Emo.} & \textbf{Aud.} & \textbf{Clarity} & \textbf{Evid.} & \textbf{Rebut.} & \textbf{Fair.} & \textbf{AVG} 
 & \textbf{Emo.} & \textbf{Aud.} & \textbf{Clarity} & \textbf{Evid.} & \textbf{Rebut.} & \textbf{Fair.} & \textbf{AVG} \\
\midrule
\textit{bert\_cs}           & 0.31 \textsuperscript{$\pm$ 0.04} & 1.00 \textsuperscript{$\pm$ 0.00} & 0.52 \textsuperscript{$\pm$ 0.11} & 0.91 \textsuperscript{$\pm$ 0.03} & 0.78 \textsuperscript{$\pm$ 0.06} & 0.54 \textsuperscript{$\pm$ 0.09} & 0.68 \textsuperscript{$\pm$ 0.02}  & 0.61 \textsuperscript{$\pm$ 0.15} & 0.99 \textsuperscript{$\pm$ 0.00} & 0.98 \textsuperscript{$\pm$ 0.01} & 0.97 \textsuperscript{$\pm$ 0.01} & 0.97 \textsuperscript{$\pm$ 0.01} & 0.99 \textsuperscript{$\pm$ 0.00} & 0.92 \textsuperscript{$\pm$ 0.03} \\
\textit{bert\_cs\_hs}       & 0.30 \textsuperscript{$\pm$ 0.04} & 1.00 \textsuperscript{$\pm$ 0.00} & 0.49 \textsuperscript{$\pm$ 0.16} & 0.91 \textsuperscript{$\pm$ 0.04} & 0.77 \textsuperscript{$\pm$ 0.07} & 0.48 \textsuperscript{$\pm$ 0.07} & 0.66 \textsuperscript{$\pm$ 0.02}  & 0.69 \textsuperscript{$\pm$ 0.13} & 0.99 \textsuperscript{$\pm$ 0.00} & 0.97 \textsuperscript{$\pm$ 0.01} & 0.97 \textsuperscript{$\pm$ 0.02} & 0.88 \textsuperscript{$\pm$ 0.19} & 0.97 \textsuperscript{$\pm$ 0.03} & 0.91 \textsuperscript{$\pm$ 0.05} \\
\textit{multi-task\_d.}     & 0.51 \textsuperscript{$\pm$ 0.00} & 1.00 \textsuperscript{$\pm$ 0.00} & 0.45 \textsuperscript{$\pm$ 0.00} & 0.97 \textsuperscript{$\pm$ 0.00} & 0.62 \textsuperscript{$\pm$ 0.00} & 0.49 \textsuperscript{$\pm$ 0.00} & 0.67 \textsuperscript{$\pm$ 0.00}  & 0.95 \textsuperscript{$\pm$ 0.00} & 1.00 \textsuperscript{$\pm$ 0.00} & 0.98 \textsuperscript{$\pm$ 0.00} & 0.96 \textsuperscript{$\pm$ 0.00} & 0.94 \textsuperscript{$\pm$ 0.00} & 0.95 \textsuperscript{$\pm$ 0.00} & 0.96 \textsuperscript{$\pm$ 0.00} \\
\textit{multi-task\_u.}     & 0.47 \textsuperscript{$\pm$ 0.00} & 1.00 \textsuperscript{$\pm$ 0.00} & 0.31 \textsuperscript{$\pm$ 0.00} & 0.97 \textsuperscript{$\pm$ 0.00} & 0.60 \textsuperscript{$\pm$ 0.00} & 0.45 \textsuperscript{$\pm$ 0.00} & 0.63 \textsuperscript{$\pm$ 0.00}  & 0.91 \textsuperscript{$\pm$ 0.00} & 1.00 \textsuperscript{$\pm$ 0.00} & 0.96 \textsuperscript{$\pm$ 0.00} & 0.94 \textsuperscript{$\pm$ 0.00} & 0.96 \textsuperscript{$\pm$ 0.00} & 0.96 \textsuperscript{$\pm$ 0.00} & 0.96 \textsuperscript{$\pm$ 0.00} \\
\textit{dependency\_m.\_3e}  & 0.45 \textsuperscript{$\pm$ 0.00} & 1.00 \textsuperscript{$\pm$ 0.00} & 0.32 \textsuperscript{$\pm$ 0.00} & 0.97 \textsuperscript{$\pm$ 0.00} & 0.60 \textsuperscript{$\pm$ 0.00} & 0.51 \textsuperscript{$\pm$ 0.00} & 0.64 \textsuperscript{$\pm$ 0.00}  & 0.86 \textsuperscript{$\pm$ 0.00} & 1.00 \textsuperscript{$\pm$ 0.00} & 0.97 \textsuperscript{$\pm$ 0.00} & 0.95 \textsuperscript{$\pm$ 0.00} & 0.96 \textsuperscript{$\pm$ 0.00} & 0.98 \textsuperscript{$\pm$ 0.00} & 0.95 \textsuperscript{$\pm$ 0.00} \\
\textit{dependency\_m.\_6e}  & 0.59 \textsuperscript{$\pm$ 0.00} & 1.00 \textsuperscript{$\pm$ 0.00} & 0.41 \textsuperscript{$\pm$ 0.00} & 0.97 \textsuperscript{$\pm$ 0.00} & 0.60 \textsuperscript{$\pm$ 0.00} & 0.61 \textsuperscript{$\pm$ 0.00} & \textbf{0.70} \textsuperscript{$\pm$ 0.00}  & 0.95 \textsuperscript{$\pm$ 0.00} & 1.00 \textsuperscript{$\pm$ 0.00} & 0.98 \textsuperscript{$\pm$ 0.00} & 0.98 \textsuperscript{$\pm$ 0.00} & 0.98 \textsuperscript{$\pm$ 0.00} & 0.99 \textsuperscript{$\pm$ 0.00} & \textbf{0.98} \textsuperscript{$\pm$ 0.00} \\
\bottomrule
\end{tabular}
}
\label{tab:results_twitter_conan_cs}
\end{table*}

\textbf{In-Domain Evaluation.} 
Table \ref{tab:results_twitter_conan_cs} reports F1 classification results of model configurations, with training and testing on the same dataset for the Twitter Dataset and for CONAN. Training and testing on the same dataset confirm that dependency-based models consistently perform best across effectiveness dimensions, with \textit{Dependency\_matrix\_6e} again achieving the best average F1 score, further supporting the benefit of modeling interrelations between dimensions in classification tasks.
We also notice that training and testing on expert-written CS (CONAN) seem easier, as it achieves higher average F1 scores, compared to training and testing on the Twitter dataset.

The results across all three evaluation settings highlight key insights into the classification of counter-speech effectiveness dimensions. While emotional appeal presents unique challenges due to its subjectivity, the other dimensions (audience adaptation, clarity, evidence, rebuttal, and fairness) also exhibit distinct patterns that provide valuable directions for model improvement.

\section{Error Analysis and Discussion}
\label{sec:error_discussion}

In this section, we discuss some examples to better understand the models' behaviors on two dimensions, Emotional Appeal (most difficult to classify) and Audience Adaptation (highest F1 scores), and we address some limitations.

\noindent \textbf{Audience Adaptation}.
It consistently achieves high F1 scores (0.99–1.00) across models, likely due to the similarity in linguistic structure and content between CS and HS. For example, consider the following HS/CS pair from CONAN:

\noindent \textbf{HS}: \textit{"Islam is not a religion that professes peace and it does not make any good men. It condemns women and encourages delinquency. Lawbreakers are not all of a certain race or religion, but only one race and religion make lawbreakers."}

\noindent \textbf{CS}: \textit{"Today, in many Muslim-majority communities, women receive equal, even higher in some places, treatment as men. For example, many women work as scientists, engineers, lawyers, etc."}
Both speeches aim to persuade a specific group by matching linguistic complexity, thus ensuring audience adaptation. The CS provides counter-examples to the HS claims, using similarly complex and detailed arguments. This makes it easier for the classifier to recognize this dimension, as both speeches maintain a similar linguistic and argumentative structure. However, this near-perfect performance could reflect data bias, suggesting that audience adaptation might be too easily detected, due to the class imbalance.
Further error analysis should investigate whether models are truly understanding adaptation strategies or simply leveraging surface-level cues.

\noindent \textbf{Emotional Appeal.}
Emotional appeal is challenging to classify due to its subjective nature and context-dependent expression. For example, consider the following counter-speeches from Twitter and CONAN:

\noindent \textbf{Twitter}: \textit{"@user this isn't funny at all. don't make jokes of the situation when rene herself is affected by it. don't get her dragged even more and delete this."}
This CS is direct and uses imperatives to express frustration and urgency, relying on internet culture and social cues like tagging someone and using second-person pronouns ("you").

\noindent \textbf{CONAN}: \textit{"I thought we'd stopped exporting our convicted criminals last century. Now you advocate exporting people without a trial or a conviction."}
This example uses sarcasm, rhetorical questioning, and historical references to evoke moral outrage and challenge the opposing argument. The emotional appeal here is more subtle and indirect, contrasting with the explicit frustration in the Twitter example.
A model trained solely on Twitter might miss these subtleties found in expert-written CS. To improve emotional appeal classification, domain adaptation or dataset augmentation could help models recognize diverse emotional expressions.

\noindent \textbf{Model Limitations.}
We use BERT-based classifiers throughout to ensure that improvements stem from our multi-task and dependency-based framework rather than the underlying model. Our goal was not to benchmark architectures but to evaluate the framework itself. For example, while Large Language Models (LLMs) are promising, and can  achieve good results, they are optimized for generation rather than classification tasks. Evaluating LLMs would also require a separate experimental design, which is beyond the scope of this paper, and we leave 
these comparisons as future work.

\section{Conclusion}
\label{sec:conclusion}

As a first contribution of this paper, we release a novel linguistic resource to the community, enriching the CONAN and Twitter dataset with manual annotations of six fine-grained dimensions of counter-speech effectiveness: Emotional Appeal, Audience Adaptation, Clarity, Evidence, Rebuttal, and Fairness. This new annotation layer provides a richer, more diverse foundation for research on CS evaluation, considering content from both expert- and user-generated sources.

As a second contribution, we introduce a framework for automatically classifying counter-speech effectiveness across these dimensions. Our approach is architecture-agnostic and compatible with a range of transformer backbones. Experimental results demonstrate that a dependency-based model, which explicitly models inter-dependencies among the dimensions, significantly outperforms standard BERT baselines, achieving 0.96 F1 score on the combined dataset. This includes strong performance on both expert- (CONAN) and user-written (Twitter) CS. In addition, our multi-task learning models also surpass BERT with 0.94 F1 score, reinforcing the hypothesis that the six dimensions are interrelated and can be jointly learned more effectively. Our cross-domain evaluation reveals substantial domain shifts between CONAN and Twitter, especially in Emotional Appeal, highlighting the challenges of generalization and the need for domain-adaptive modeling techniques. Despite this, our framework demonstrates robust improvements over BERT baselines across domains, suggesting its effectiveness can generalize beyond specific model architectures.

Finally, we conduct a detailed linguistic analysis of model performance across dimensions, revealing which aspects of CS effectiveness remain the most difficult to capture. This analysis can inform the design of future CS evaluation systems and guide annotation efforts where additional human oversight is most valuable.


\section*{Acknowledgment}
This work has  been partially supported by the French government, through the 3IA Côte d’Azur investments in the project managed by the National Research Agency (ANR) with the reference number ANR-23-IACL-0001.

\bibliographystyle{IEEEtran}
\bibliography{main}

\end{document}